\RequirePackage{fix-cm}
\documentclass[twocolumn]{svjour3}          
\smartqed  
\usepackage{graphicx}
\begin{document}

\title{Improving a neural network model by explanation-guided training for glioma classification based on MRI data
}


\author{František Šefčík         
\and
        Wanda Benesova
}


\institute{ F. Šefčík \at
             Slovak University of Technology in Bratislava, Bratislava 84216, Slovakia \\
            \email{xsefcik@stuba.sk}
           \and
           W. Benesova \at
           Slovak University of Technology in Bratislava, Bratislava 84216, Slovakia, ORCID 0000-0001-6929-9694 \\
            \email{vanda\_benesova@stuba.sk}
}

\date{Received: date / Accepted: date}

\maketitle

\begin{abstract}

In recent years, the artificial intelligence (AI) systems have come to the forefront. These systems, mostly based on Deep learning (DL), achieve the excellent results in areas such as image processing, natural language processing or speech recognition. Despite the statistically high accuracy of deep learning models, their output is often a decision of "black box". Thus, Interpretability methods have become a popular way to gain insight into a decision making process of deep learning models. Explanation of a deep learning model is desirable in medical domain since the experts have to justify their judgments to the patient.

In this work, we proposed method for explanation-guided training that uses a Layer-wise relevance propagation (LRP) technique to force model to focus only on the relevant part of the image.We experimentally verified our method on a convolutional neural network (CNN) model for low-grade and high-grade glioma classification problem. Our experiments show promising results in a way to use interpretation techniques in the model training process.

\end{abstract}

\section{Introduction}
\label{introduction}
In recent years, we have been increasingly encountering the term Deep learning (DL) as a group of Machine learning (ML) approaches. DL methods such as deep neural networks gain success in a wide range of applications, for example in the image classification, speech recognition or natural language processing. These techniques achieve high levels of performance in many complex tasks due to their learning, reasoning, and adaptation capabilities while being able to outperform human accuracy. These features make them pivotal for the future progress of human society.

Despite the statistically high accuracy of deep learning methods, their output is often "black-box" decision. The black-boxness is the result of models' complexity e.g., hundreds of layers and millions of parameters. The opposite of black-boxness is transparency which is a direct understanding of the mechanism by which the model works. \cite{mythosXAI}  In recent years, methods of interpretability have become popular in way to obtain insight into the decision making of DL models. Methods of interpretability and explainability have become popular in the fields such as self-driving cars or medicine, where the relevance of the features underlying the model must be guaranteed because these decisions can affect human lives. \cite{peeking}

In the medical domain, Explainable artificial intelligence (XAI) is extremely important for the further integration of deep learning methods. Medical experts should get an opportunity to know how and why models decisions were made by models. \cite{medicalXAI} This knowledge is essential in using AI in medicine to justify medical decisions to the patient. Transparent algorithms will increase the trust of experts in new emerging systems, along with patients' trust.

In this way, we trained a CNN model to classify LGG (low-grade glioma) and HGG (high-grade glioma) disease from 2D MRI data. To evaluate and better understand model's behavior, we implemented a method for detection of similar cases and their explanation by visualization. We used an inner representation of CNN with the analysis of neuron activations on hidden layers in forward pass of the network to predict similar atlas images to an input image. On the basis of the inner representation and neuron activations on hidden layers of the CNN model we proposed an explanation method based on BiLRP technique\cite{Eberle_2020} to show common features between an input and sample atlas images. This explanation provides higher-order explanations that extend conventional visualization techniques.
 
Using this method, we evaluated that the model does not work correctly with a relatively high accuracy. Explanation by visualization showed that our model for brain tumor classification learnt incorrect features for decision making. The model should pay much more attention to the regions with tumors. Based on these observations, we proposed an explanation-guided training method to improve the model's prediction. Our method exploits the additional information related to the inputs that can help the model with better understanding of the problem. In our solution, we forced the model to focus more on the tumor region in 2D MRI brain slices. The proposed method has potential in training models when we want to prefer some well-known features from the input, but we do not want to restrict the model only to these features. The main steps of our work are described as follows:

\begin{itemize}
  \item We implemented a method for explanation of the model by predicting similar cases to input image and explaining it by visualizing similar features between these images.
  \item We used our explanation method to derive an explanation of the trained model to reveal the pitfalls of the predictions.
  \item We propose a method of explanation-guided training to prioritize relevant features.
  \item We performed experiments to show the results of the proposed method on trained model for glioma disease classification.
  \item We outline promising applications of the proposed method in medicine and other fields.
\end{itemize}

Hence, the main contribution of this article can be summarized as follows:
We introduce a novel method of explanation-guided training of DNN to prioritize relevant features in the input during the training phase. This method has been evaluated for medical application.

\section{Related Work}\label{related}
Explainability is one of the main obstacles AI currently faces in practical implementation. The needs for XAI have been commonly described and broken down by authors in 
\cite{peeking,gunning2017explainable,samek2018explainable} into four main motivations: reasoning about decisions for people who will be affected by the decisions of these systems; control over the system to prevent things from going wrong; improving models; and discovery to uncover patterns that humans cannot capture.

To achieve these goals, many XAI methods have emerged with different approaches for a spectrum of problems. XAI methods can be divided into two main groups as \textit{transparent} models and \textit{post-hoc} explainability. \textbf{Transparent models} are models interpretable by their design. The purpose of the method is to design transparent models that are self-explanatory. Post-hoc explainability is explaining of the "black-box" model. It can be understood as the application of external interpretation techniques to the "black-box". The vast group of Post-hoc techniques has been divided into several subgroups in \cite{arrieta2020explainable} as: explanation by simplification; feature relevance explanation; explanation by visualizing hidden abstractions; and architecture modification.

The XAI methods adopted to explain DNNs, specifically on CNN, can be divided into two main categories. Methods in the first category seek to understand the decision process by mapping the output back to the input to determine which part of the input was discriminative for the output. The second category methods try to delve inside the network and interpret how the intermediate layers see the external world. These methods do not focus on any specific input, but rather on a general representation of the model. \cite{arrieta2020explainable}

One of the most straightforward methods interpreting the DNN prediction is \textit{Perturbation based methods}. This broad category of techniques perturbs intensities in the input image and observes the changes in prediction probabilities. The main idea is that the pixels that contribute maximally to the prediction will reduce the prediction probability if they change.

\textit{Occlusion sensitivity} was presented in the paper\\Zeiler et al. \cite{zeiler2014visualizing}. In this approach, they occluded a part of the image with a gray window, and they were sliding this window cross the whole image. For each window position, prediction probability is monitored and captured by color range. Finally, we can see heatmap with the areas that have the greatest impact on a particular prediction. This idea was developed also in another approach named \textit{Randomized Input Sampling for Explanation (RICE)}. \cite{rice} They obtain the importance map by probing the model with randomly masked versions of the input image instead of a sliding window over input image. \textit{Local Interpretable Model-agnostic Explanations (LIME)} \cite{ribeiro2016should} is another type of \textit{Perturbation methods} that build surrogate model around black-box predictions to explain it. LIME train local surrogate model to approximate prediction of complex models instead of training global surrogate model. Perturbation-based methods work well for explaining decisions but suffer from computational cost and instability to surprising artifacts.

Methods based on gradient offer another way to interpret models. These methods involve a single forward and backward pass through the network, in addition to multiple forward passes in Perturbation-based methods. Also, gradient-based methods are computationally more efficient and stable in artifacts compared to perturbation-based methods.

One of the methods from this category is \textit{Sensitivity analysis} \cite{sensitivity} and it assumes that the most relevant input features are those to which the output is most sensitive. It attempts to explain the prediction based on the local evaluation of the gradient of the model. Sensitivity analysis mathematically quantifies the relevance of input variables $R_i=\left(\frac{\partial F(x)}{\partial x_i}\right)^2 $
This technique is easy to implement for neural networks because the gradient can be computed using backpropagation. 

The purpose of applying sensitivity analysis is usually not to explain the relationships found. Instead, it is generally used to test the stability and credibility of models, as well as a tool to remove some unnecessary input attributes or as a starting point for more complex explanatory techniques such as decomposition. \cite{samek2018explainable}

There is some more modification of gradient's methods as \textit{Integrated Gradient} \cite{integrated_gradient}, \textit{Gradient*Input} \cite{simonyan2013deep}, \textit{Guided Backprop} \cite{springenberg2014striving} or \textit{Grad-CAM} \cite{selvaraju2017grad}. These algorithms differ in the way how gradients are modified in a backward pass. However, these techniques must deal with the problems in propagating gradient through non-linear layers.

Another approach to explain the prediction of Deep neural networks is the explicit use of their graph structure. These methods use the following procedure: starting at the network output, then prediction is mapping in the reverse direction of the graph, so we always map the prediction to lower layers until we reach the input of the network. \cite{montavon2018methods} 

\textit{Layer-wise relevance propagation (LRP)} \cite{bach2015pixel}, technique which was developed specifically for explaining neural networks, is based on this principle. Thus, each neuron in NN receives a fraction of the output and further distributes it to its predecessors in equal amounts until the input is reached.\cite{lrp}

In the paper from Papernot et al. \cite{papernot2018deep}, an approach that is completely different from previous methods is proposed. They apply \textit{K-Nearest Neighbour} algorithm to the representation of data learned by layers in CNN to understand model failures. The test input is compared with neighboring training points by distance in representation learned in hidden layers of CNN. Then, the confidence of the model is taken as the homogeneity among neighbour labels. We can detect adversarial examples or protect model from examples that are outside of model understanding.

\textit{BiLRP} \cite{Eberle_2020} is method based on second-order explanation. This method was introduced to explain similarity between two inputs of model. The method was inspired by LRP to bring robustness for explaining of dot product similarities. BiLRP performs a second-order \textit{deep Taylor decomposition} \cite{montavon2017taylor} of the similarity score, which enable to examine common features that contribute to similarity on any hidden layer of DNN.

Explanation techniques are usually used to explain decisions of trained model. Authors of the recent work \cite{sun2020explanation} presented explanation-guided training for task of cross-domain few-shot classification. They developed a model-agnostic explanation-guided training strategy\\based on LRP method that dynamically finds and emphasizes the features which are important for the predictions. Their work shows application of explanation for training phase which effectively improves the model generalization.

\section{Proposed method}\label{method}

\paragraph{Problem description}

Before presenting our method for explaining and improving model prediction by interpretation technique, we firstly introduce the task of LGG (low-grade glioma) vs. HGG (high-grade glioma) diagnosis classification problem which is a combination of tumor region segmentation and patient survival prediction. In general, LGG cells do not attack normal neighboring cells, while HGG cells attack on their adjacent cells \cite{wu2018grading}. Accurate classification of gliomas is therefore an important requirement because the type of glioma has an impact on the patient's overall survival. This problem has also been discussed in other papers \cite{cho2017classification},\cite{akkus2017predicting},\cite{mehrotra2020transfer}.

\paragraph{Dataset}

To train our method, we chose BraTS dataset \cite{menze2014multimodal}. Dataset consists of multi-modal MRI scans of glioblastoma (GBM/HGG) and lower grade glioma (LGG) with pathologically confirmed diagnosis. Segmented annotations are available for each volume.

\paragraph{Model and training}

\begin{figure}[h]
    \centering
    \includegraphics[width=1\linewidth]{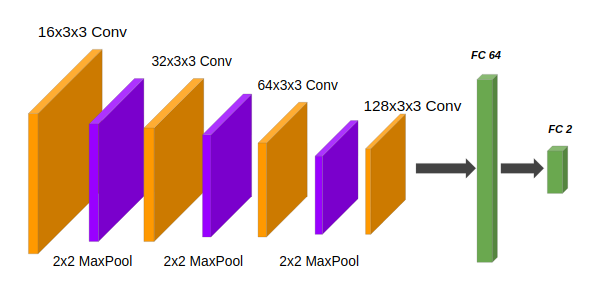}
    \caption{Network architecture.}
    \label{fig:model}
\end{figure}

In the preprocessing phase, we used three MRI sequences (T1-weighted and T2-weighted and FLAIR) to generate three-channel 2D tumor slices from MRI brain volumes. For each class, we extracted a different numbers of slices with largest area of tumor region to balance the dataset. In the next step, we trained model of Convolutional Neural Network with prepared 2D tumor slices for classification of LGG and HGG tumors. Architecture of the network is built from four 2D convolution layers with ReLu activation function followed by Max pooling layers, end up with two fully connected layers. Detailed description of network's layers is showed in Table \ref{fig:model}

To avoid fast overfitting, we applied a random rigid augmentation with rotation, horizontal and vertical flip transformations. Neuron dropout was also applied after first and last max-pooling layer. Adam's optimizer with Sparse Categorical Crossentropy loss function was used as the optimization function.

\subsection{Detection of similar cases and their explanation by visualization}\label{explanation}

\begin{figure}[h]
    \centering
    \includegraphics[width=1\linewidth]{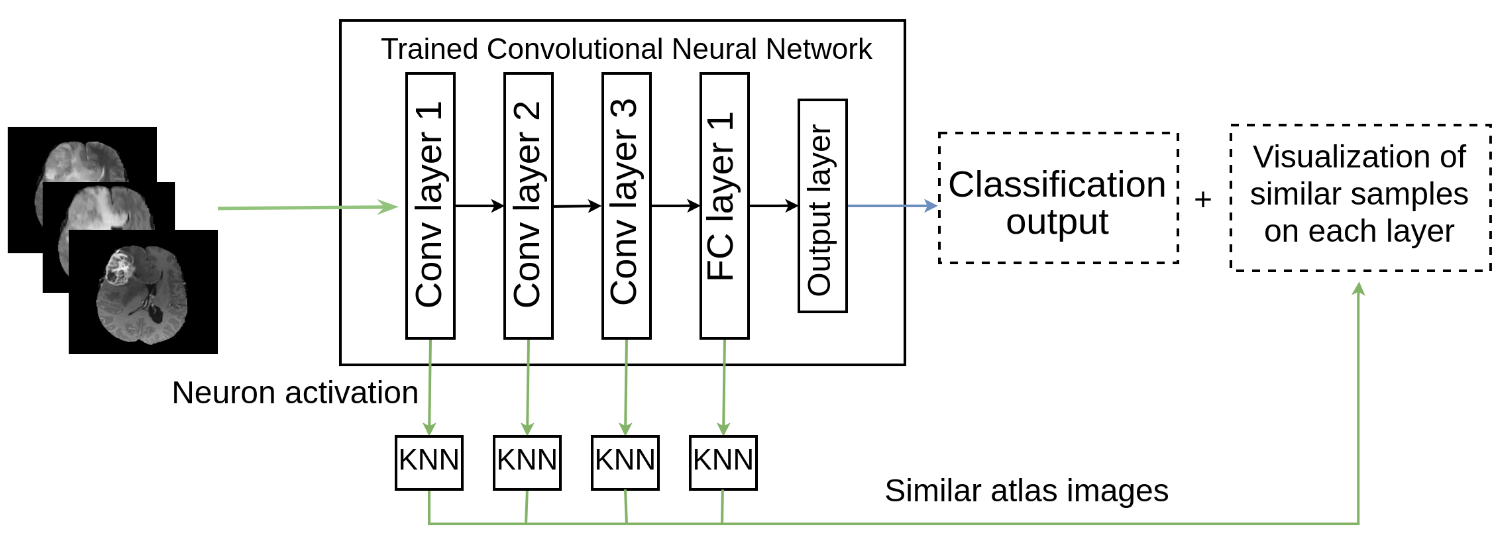}
    \caption{Overview designed tool for similar cases detection and their explanation by visualization.}
    \label{fig:similar_cases}
\end{figure}

To explain our trained model, we implemented a method for detection of similar cases and their explanation by visualization showed in Figure \ref{fig:similar_cases}. This method is not an essential part of the paper, but forms the basis for the proposed method in the following section \ref{improvements}. The method is based on prediction similar atlas images based on their features in hidden layers, inspired by \textit{Deep k-nearest neighbors} \cite{papernot2018deep}. The method uses KNN classifier to find the most similar images to the input from neurons activations on every hidden layer. The output of the method is used to allow domain experts to see pairs of similar cases in network predictions.

Here, domain experts can ask the question: What led to the prediction? The answer to this question can be to use any explanatory technique to explain what the network sees in the two images. Then the experts can address how the images are related to each other. Here we proposed to use any of higher-order explanations to show domain experts which features in the images are commonly similar. We can also point out that this technique can be the starting point for \textit{Contrastive explanations} \cite{MILLER20191}.

To bring these higher-order explanations to MRI data classification, we introduce using of BiLRP \cite{Eberle_2020} technique. In our method for detecting similar cases, similar atlas images to the input image are visualized. Therefore, we can represent the classification of hidden layer features using KNN as a similarity model, and the BiLRP method suitable for this type of problem. We applied BiLRP to the features of all selected hidden layers. We consider the selection of features on the hidden layer as crucial for the explanation and their subsequent interpretation for the user.

\begin{figure}[h]
    \centering
    \includegraphics[width=1\linewidth]{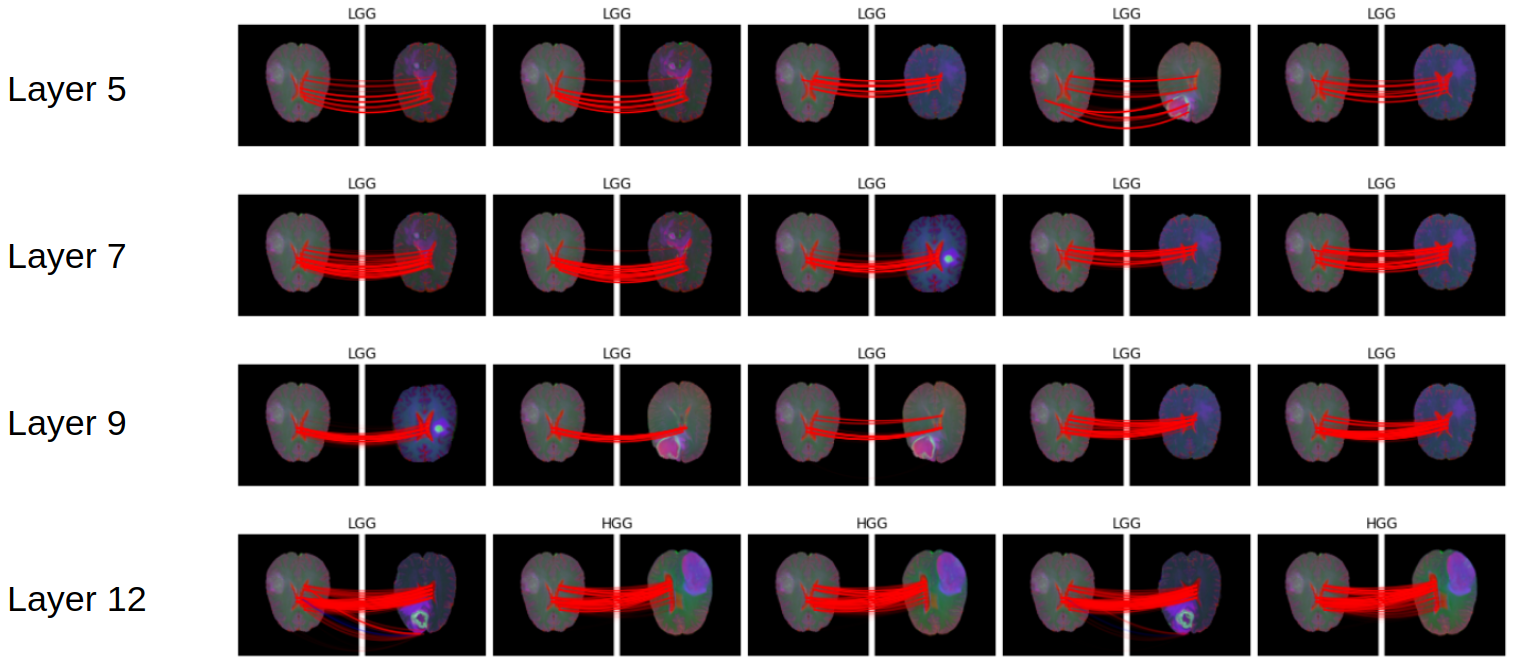}
    \caption{Explanation of similar features between atlas images and input image by BiLRP. Red lines between images show their similar features relevant for prediction.}
    \label{fig:bilrp_pair}
\end{figure}

Figure \ref{fig:bilrp_pair} shows the result of our method for similar cases detection and their visualization by BiLRP explanation. An important conclusion follows from these explanations. The model should point to areas of brain tumors; instead, the model focuses on areas that are not primarily related to brain tumors. This suggests that the model did not learn the correct information which we would consider the most relevant to the model, despite the model's reasonably good accuracy. These observations led us to develop a method to address this problem.

\subsection{Novel method of explanation-guided training}\label{improvements}

\begin{figure*}[h!]
  \includegraphics[width=0.9\textwidth]{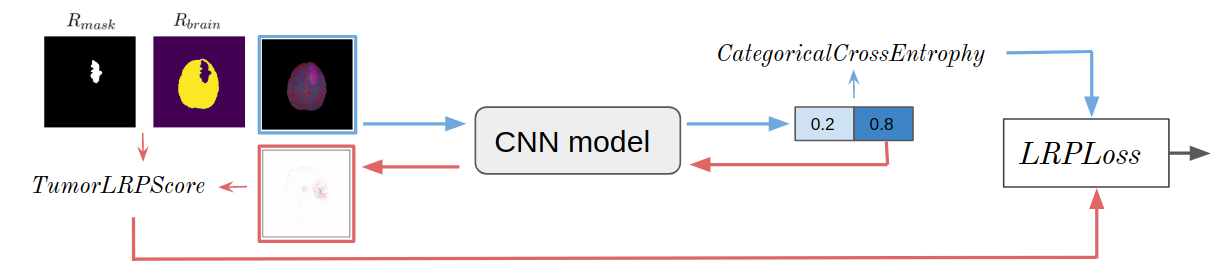}
\caption{Novel method of explanation-guided training. Blue path represents the classic training phase. Red path represents enhancement of original training process.}
\label{fig:lrp_loss}
\end{figure*}

The proposed method introduces a novel way of training the neural network. It brings type of explanation-guided training where we can use complementary information in form of segmentation mask to force model to focus only on relevant part of the image. The method does not require changes in architecture of the model, but it modifies loss function in training phase of the model. One training phase involves two following steps illustrated in figure \ref{fig:lrp_loss} as blue and red paths. For each training iteration: \textbf{1st step (blue)} The input passes forward through the network to obtain prediction value $pred$ as the output. \textbf{2nd step (red)} involves interpretation method LRP to obtain relevance score $R$ from output $pred$. LRP is initiated from neuron corresponding to the true label $y$.  The $R$ score tells how much each pixel of the input image contributes to the prediction. With segmentation mask of the relevant region from input image we can observe if relevant pixels with high score $R$ are present in the region. Thus, we can quantify how much attention the network gives to the relevant region. We used this information to modify training phase and to design new loss function.

We applied proposed method to the aforementioned task of tumor disease classification. In addition to evaluating how accurately the model classifies between the LGG and HGG classes, the newly proposed loss function also uses an LRP interpretation technique to penalize the model if it does not target the tumor region during prediction. The loss function is named $LRPLoss$ and can by written as:

\begin{equation} 
LRPLoss = \frac{CategoricalCrossEntrophy(pred, y)}{TumorLRPScore(pred, x_{seg})}  
\end{equation}

where $CategoricalCrossEntrophy$ is calculated from $pred$ score as output of network and $y$ as true label. This loss value is penalized with $TumorLRPScore$ which describes how much attention is paid to the tumor region. When the score is high then the most relevant pixels are situated in the relevant region of tumor. The score is calculated as: 

\begin{equation} 
TumorLRPScore=\frac{R_{mask}}{R_{mask} + R_{brain}} 
\end{equation}

where $R_{mask}$ is sum of LRP relevances inside of the tumor mask while $R_{brain}$ is sum of LRP relevances of the whole brain mask without tumor region. Different types of LRP method can be used. In some of them, the relevances can acquire positive and negative values, then we count only positive values. We also use version of $TumorLRPScore$ where both its components are normalized by size of their regions. After the observation, we learnt that the first version of scores describes the property in better manner. Subsequently, we were training the models with designed loss function.

In conclusion, we have designed novel loss function, which is used to modify the training phase. In ordinary training, we use images with their labels. In proposed training, we need images, segmentation masks of tumors and labels.

\section{Experiments and results}

To evaluate our method, we proposed two experiments. In first experiment, we applied proposed method to our trained model where we compare different levels of penalization in $LRPLoss$. In second experiment, we applied our method to training of state-of-the-art solution from Subhas, B. et. al.\cite{banerjee2019deep} Their solution is comparable with our model with similar data pipeline where  slices  are  extracted  from  volume  as  2D  images. They achieved Accuracy 0.86, Specificity 0.70 and Sensitivity 0.92.

To train and evaluate the model, we selected the BraTS dataset \cite{menze2014multimodal}. The dataset consists of multimodal MRI images of glioblastoma (GBM/HGG) and lower-grade glioma (LGG) with pathologically confirmed diagnosis. Data are provided in train, validation, and test sets, but ground truth data are only available for train data. The training set consists of 260 HGG samples and 76 LGG samples. All BraTS multimodal scans are available in multiple clinical protocols as T1- and postcontrast T1-weighted, T2-weighted (T2) and T2 Fluid Attenuated Inversion Recovery (FLAIR) volumes. Segmented annotations are available for each volume.

 
 


In the evaluation of the proposed method, we looked at several factors which influence accuracy of the model. We evaluate performance of the model with metrics as Accuracy and F1 weighted score. Alongside the evaluation of the model performance, we focused on evaluating how the proposed loss function influenced model in the way of choosing right features in a decision process. To quantify this metric, we calculate mean of \\$TumorLRPScore$ for both classes (LGG and HGG) in each experiment. The score was defined in the\ref{improvements} section and generally tells how model targets tumor region.

\paragraph{Experiment 1}

In table \ref{tab1} bellow, we can see the results from experiment in which we trained four same models but the loss function was modified. The first model was trained with the $Original$ Categorical cross entrophy loss function, this model is our baseline model without any changes in the training phase. The rest of the models are trained with introduced loss function where TumorLRPScore is to power of 1, 2, and 3 as $Penalization 1/2/3$. By increasing of exponents, we test penalization of the loss value with a higher power. This can be interpreted as defining the level of prioritization for features in the relevant area.  

\begin{table}[h!]
\caption{Experimental results of proposed explanation-guided method. LGG and HGG score is average $TumorLRPScore$ for each class.}\label{tab1}
\begin{tabular}{ |l|l|l|l|l| }
 \hline
 {\scriptsize Loss function} & {\scriptsize Accuracy} & {\scriptsize F1 score} & {\scriptsize LGG score} & {\scriptsize HGG score}\\
 \hline
 {\scriptsize $Original$} & 0.72    & 0.71 &   0.26 & 0.22\\
 \hline
  {\scriptsize $Penalization 1$} & 0.79 & 0.79  & 0.36 & 0.36\\
 \hline
 {\scriptsize $Penalization 2$} & 0.72 &  0.70 &  0.43 & 0.61\\
 \hline
  {\scriptsize $Penalization 3$} & 0.74 & 0.74 &  0.46 & 0.65\\
 \hline
\end{tabular}
\end{table}

\begin{itemize}
  \setlength\itemsep{1em}
  \item \textit{Original} = {\scriptsize $CategoricalCrossEntrophy$}
  \item $Penalization 1 = \frac{CategoricalCrossEntrophy}{TumorLRPScore}$
  \item $Penalization 2 = \frac{CategoricalCrossEntrophy}{TumorLRPScore^2}$
  \item $Penalization 3 = \frac{CategoricalCrossEntrophy}{TumorLRPScore^3}$
\end{itemize}

The results showed a slight improvement in the accuracy of the models with the new loss function. An important observation is in the columns with the LRP score, which describes the attention of the model towards the tumor region. The results showed an increase in scores for both classes, with the gain being more pronounced for the HGG class. This could be an indication that our method helps the model better detect the features of the HGG class, which is more aggressive. Another observation talks about the strength of the penalization, we see that higher power affects the fact how much attention the model pays to the relevant regions in the input images. 

The presented results can also be interpreted in the following figure \ref{fig:lrp_loss2}. Each row shows predictions of four aforementioned models sequentially from the left. On the left, is the input image with true label and the corresponding tumor segmentation mask followed by four LRP heatmaps, each for one of the trained models. Each heatmap is accompanied by pair of LRP scores and predicted labels drawn above.

\begin{figure}[h]
    \centering
    \includegraphics[width=1
    \linewidth]{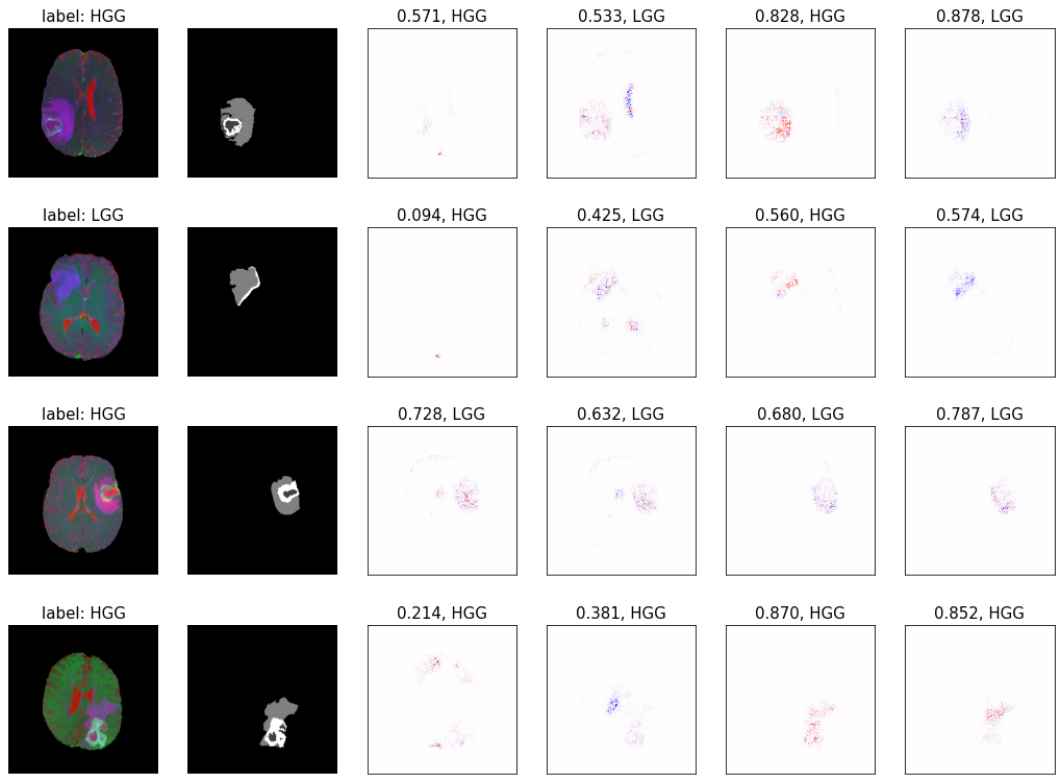}
    \caption[Predictions of four different models explained by LRP heatmap.]{Predictions of four different models explained by LRP heatmap.}
    \label{fig:lrp_loss2}
\end{figure}

The prediction examples in \ref{fig:lrp_loss2} figure represent the visual result of the proposed method application. In the heatmaps of the original model without changes to the loss function, we see that the LRP relevance is scattered throughout the brain. On the other hand, in the heatmaps where our method was used, we see an improvement in the intensity of the relevancies in the tumor regions, and also these relevancies are no longer scattered throughout the image. This is practical projection of our proposed method.

\paragraph{Experiment 2}

In the second experiment, we trained the state-of-the-art model from Subhas, B. et. al. al.\cite{banerjee2019deep} with the proposed explanation-driven training and our $LRPLoss$. Here, we observed the changes in the training process caused by our method compared to conventional training.

The model was trained in 20 iterations and in each iteration, we captured the accuracy of the model on the validation data and the Mask LRP score. In the following figure \ref{fig:training}, we see two plots describing the accuracy and Mask LRP scores for the model with the conventional loss function and our LRPLoss function configured as $Penalization1$ from the previous experiment.

\begin{figure}[h]
    \centering
    \includegraphics[width=1
    \linewidth]{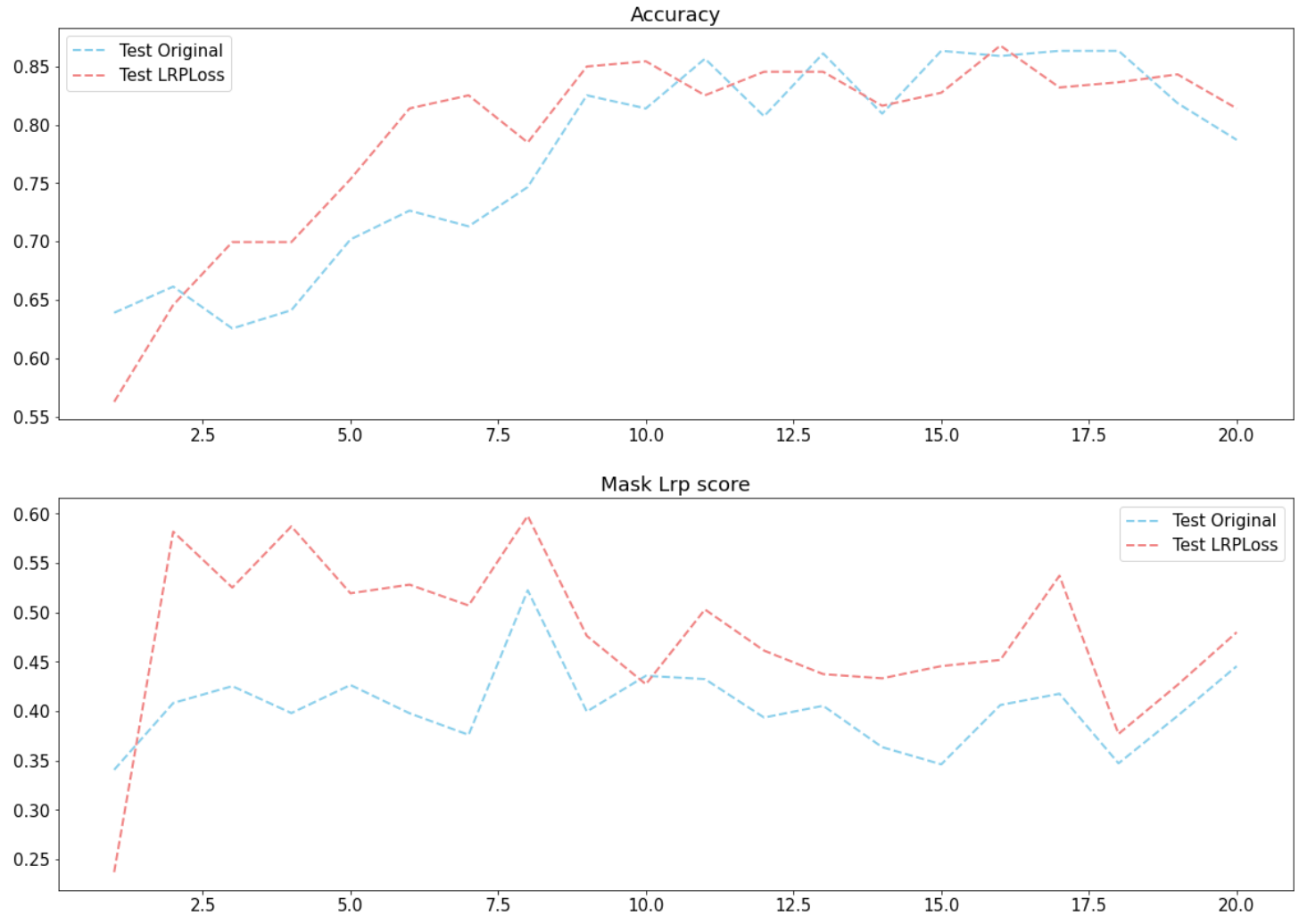}
    \caption{Training iterations of conventional loss and proposed LRPLoss function. The first graph shows Accuracy and the second one shows Mask LRP score of the models obtained from validation data. }
    \label{fig:training}
\end{figure}

From the graphs, we can see the impact of the proposed method. The main difference was gained in the first iterations of the training process, where our method helps the model achieve better accuracy in the earlier iterations. In later iterations, both models converged to similar accuracy. This behaviour can be explained as an effect of our method where the $LRPLoss$ function helps the model recognize faster the relevant features in the input image. This observation is supported by the evolution of the LRP mask score. In the first iterations, the score was significantly higher for the proposed method. Despite the decrease in scores in later iterations, the average Mask LRP score is maintained higher compared to the conventional training. We justify the convergence of accuracy and the decrease of Mask LRP scores as the dominance of $cross entropy$ loss over penalization in later iterations, because when the model has enough information from the relevant part of the image, penalization becomes less important for prediction.

\begin{figure}[h]
    \centering
    \includegraphics[width=1
    \linewidth]{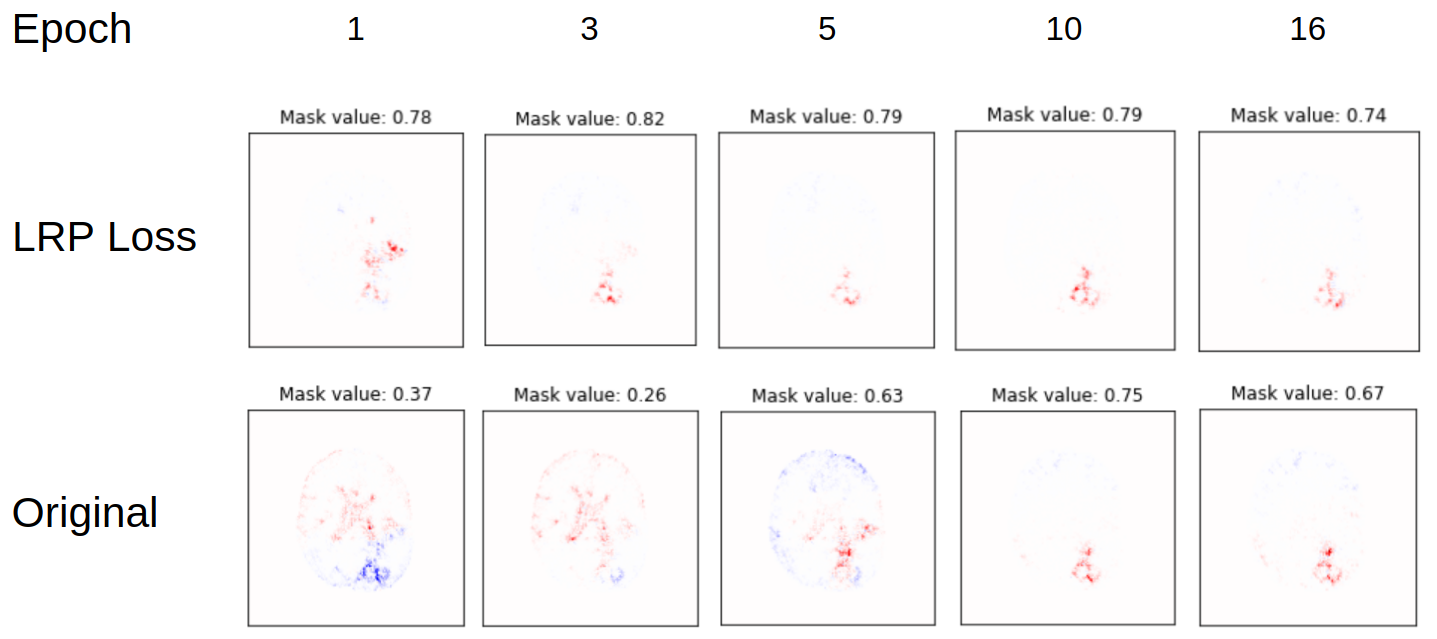}
    \caption{Explanation of prediction by LRP technique across training iterations.}
    \label{fig:training_images}
\end{figure}

The Figure \ref{fig:training_images} shows the comparison of models during training phase. The images show the relevant pixels obtained by LRP technique during the training iterations. We can see practical impact of the proposed method where we can see that our method helps the model reveal relevant part of the input image much faster then does the conventional training.

\section{Conclusion and future work}

We have proposed a method in which the interpretation technique can help train the model in case we have any additional information related to the input data. We have shown that the LRP technique can be used not only to visualize the model's decisions, but also to correct the model to make appropriate decisions. The method aims to force the model to pay more attention to any features in the input image if we required it. This method presented the loss function enhanced by LRP technique, but it is not limited only to one interpretation technique.

We applied the proposed method to the CNN model for the classification of LGG vs. HGG brain tumor disease from MRI data. We also see the perspective in application and research work in other domains. The proposed explanation-guided training can be used when we want to prefer some well-known features from the input, but we do not want to limit the model only to those features by segmentation or those areas that are scattered in the image. The method can be also used to train models of neural networks faster. Another area can be attention networks or object detection. In medical domain, we see the perspective of the proposed method in the problem of Alzheimer's disease where attention is paid to different parts of the brain depending on their features.


%
%

\bibliographystyle{spmpsci}      

\bibliography{assets/template.bib}

%
%

\end{document}